\documentclass[11pt,a4paper]{article}

\usepackage[a4paper,margin=0.9in,headheight=14pt]{geometry}
\usepackage[T1]{fontenc}
\usepackage[utf8]{inputenc}
\usepackage{lmodern}
\usepackage{microtype}
\usepackage{amsmath,amssymb}
\usepackage{graphicx}
\usepackage{booktabs}
\usepackage{array}
\usepackage{tabularx}
\usepackage{longtable}
\usepackage{enumitem}
\usepackage{xcolor}
\usepackage{hyperref}
\usepackage{url}
\usepackage{titlesec}
\usepackage{caption}
\usepackage{float}
\usepackage{tikz}
\usepackage[most]{tcolorbox}
\usepackage{fancyhdr}
\usepackage{lastpage}
\usepackage{parskip}
\usepackage{needspace}
\usetikzlibrary{arrows.meta,positioning}

\definecolor{admitblue}{HTML}{163A5F}
\definecolor{admitlight}{HTML}{EEF4F8}
\definecolor{admitgray}{HTML}{5D6670}
\definecolor{admitgreen}{HTML}{2E6E55}
\definecolor{admitred}{HTML}{9D3B3B}
    
\hypersetup{
    colorlinks=true,
    linkcolor=black,
    citecolor=black,
    urlcolor=black,
    pdftitle={ADMITBench: A Safety-Governed Reference Framework for Evaluating the Admissibility of Industrial LLM Advisories},
    pdfauthor={Refiant, Inc. and Imperial College London},
    pdfsubject={ADMITBench Technical White Paper}
}

\titleformat{\section}{\Large\bfseries\color{black}}{\thesection}{0.65em}{}
\titleformat{\subsection}{\large\bfseries\color{black}}{\thesubsection}{0.55em}{}
\titleformat{\subsubsection}{\normalsize\bfseries}{\thesubsubsection}{0.5em}{}
\setlist[itemize]{leftmargin=1.4em,itemsep=0.25em,topsep=0.35em}
\setlist[enumerate]{leftmargin=1.6em,itemsep=0.3em,topsep=0.35em}
\newcommand{\ib}{ADMITBench}
\newcommand{\claim}[1]{\vspace{0.35em}\noindent\textbf{\color{black}#1}}
\newcolumntype{Y}{>{\raggedright\arraybackslash}X}

\newtcolorbox{keybox}[1][]{
    colback=white,
    colframe=black,
    coltitle=white,
    boxrule=0.7pt,
    arc=1.5mm,
    left=3mm,right=3mm,top=2mm,bottom=2mm,
    title=#1,
    fonttitle=\bfseries
}

\newtcolorbox{warningbox}[1][]{
    colback=white,
    colframe=black,
    coltitle=white,
    boxrule=0.7pt,
    arc=1.5mm,
    left=3mm,right=3mm,top=2mm,bottom=2mm,
    title=#1,
    fonttitle=\bfseries
}

\begin{document}

\begin{titlepage}
\thispagestyle{empty}
\vspace*{1.2cm}

\begin{center}
{\Huge\bfseries\color{black} ADMITBench\par}
\vspace{0.3cm}
{\Large\bfseries A Safety-Governed Reference Framework\par}
\vspace{0.35cm}
{\LARGE\bfseries Evaluating the Admissibility of\par}
{\LARGE\bfseries Industrial LLM Advisories\par}
\end{center}

\vspace{1.2cm}
\begin{keybox}[Purpose of this document]
This white paper presents ADMITBench, a reference framework for evaluating industrial LLM advisories at the level of the proposed action. The framework implements a versioned, safety-governed evaluation contract that checks whether a recommendation is supported by the available evidence, permitted under the stated authority and procedure, and acceptable under the plant-specific consequence checks encoded in the selected evaluation profile. In this report, \emph{safety-governed} means that eligibility is determined through explicit, non-compensatory checks derived from a versioned plant profile; it does not mean that the evaluator, model, or plant has been safety-certified. Release 0.1.0 is a public reference implementation for technical and research evaluation, not an authorisation for physical execution.
\end{keybox}

\vfill

\begin{tabularx}{\textwidth}{@{}>{\bfseries}p{0.28\textwidth}Y@{}}
Prepared by & Yash Misra\textsuperscript{1}, Javal Vyas\textsuperscript{2}, Siddharth Gutta\textsuperscript{1}, and Mehmet Mercangöz\textsuperscript{2} \\
Affiliations & \textsuperscript{1}Refiant, Inc.; \textsuperscript{2}Imperial College London \\
Contact & \texttt{team@refiant.ai} \\
Document version & ADMITBench 0.1.0 / White Paper Revision 1.1 \\
Status & Technical white paper draft \\
Date & August 2026 \\
Distribution & Public technical white paper draft \\
\end{tabularx}

\vspace{1.0cm}
{\small\color{black}
Suggested citation: Misra, Y., Vyas, J., Gutta, S., and Mercangöz, M. \emph{ADMITBench: A Safety-Governed Reference Framework for Evaluating the Admissibility of Industrial LLM Advisories}. Technical White Paper, 2026.
}
\end{titlepage}
\setcounter{page}{2}

\section*{Document Control}
\addcontentsline{toc}{section}{Document Control}

\begin{tabularx}{\textwidth}{@{}>{\bfseries\raggedright\arraybackslash}p{0.25\textwidth}>{\raggedright\arraybackslash}p{0.22\textwidth}Y@{}}
\toprule
Field & Current value & Interpretation \\
\midrule
Repository release & 0.1.0 & Identifies the public code and cartridge format described in this report. \\
White paper revision & 1.1 & Reconciles the framework scope, tier semantics, non-compensatory eligibility rule, reporting requirements, and cartridge layout with repository release 0.1.0. \\
Implemented cartridges & \texttt{cstr\_alpha}; \texttt{column\_alpha} & Two shipped profiles covering a CSTR and a distillation-column setting. \\
Implementation status & Public reference implementation & The report documents the implemented evaluator and released plant profiles. \\
Future portability target & Tennessee Eastman process & Proposed test of whether the profile abstraction extends to plant-wide interactions. \\
\bottomrule
\end{tabularx}

\vspace{0.8em}
\begin{warningbox}[Important use limitation]
No score, tier verdict, simulation result, or benchmark comparison produced by \ib\ should be interpreted as functional-safety certification, site acceptance, or permission to deploy an autonomous controller. Site-specific engineering review, management of change, functional-safety assessment, operator training, and accountable human approval remain necessary.
\end{warningbox}

\tableofcontents
\clearpage

\section{Executive Summary}

\textbf{In industrial operations, the answer is not the unit of safety; the proposed action is.} Industrial LLM systems may influence the actions considered by operators during faults and abnormal situations. Evaluating whether a system identifies the correct fault is therefore insufficient: a correct diagnosis may still be followed by an action that conflicts with a procedure, exceeds delegated authority, relies on invalid evidence, or produces an unacceptable physical consequence.

\ib\ is a reference framework for evaluating this diagnosis-to-action gap. It represents each response as a structured action record and evaluates that record against a versioned plant profile. The framework implements a safety-governed evaluation contract covering record integrity, evidence and state validity, hazard understanding, action admissibility, physical consequence verification, eligible-action utility, and audit traceability. Plant-specific material is stored in versioned evaluation profiles, called \emph{cartridges} in the repository, while the common evaluation logic remains separate.

The contribution of release 0.1.0 is deliberately bounded: it provides a common action-record format, a non-compensatory evaluation hierarchy, and two implemented plant profiles for comparing the admissibility of proposed industrial actions. It does not yet establish broad industrial generality, independent profile portability, human-factors adequacy, or deployment readiness.

\begin{keybox}[Main design choices]
\begin{itemize}
    \item The unit of evaluation is the proposed action record rather than the natural-language answer alone.
    \item Diagnosis quality and action admissibility are reported separately.
    \item T0--T4 checks are inspectable and replayable for a fixed action record, evaluator version, plant profile, and simulator configuration; the engineering judgements encoded in the profile remain human-authored and require review.
    \item ``Verify'', ``hold'', and ``escalate'' are treated as valid actions when evidence is degraded or authority is insufficient.
    \item A record that fails any hard gate in T0--T4 is ineligible, receives \texttt{aggregate=None}, and is excluded from utility ranking.
\end{itemize}
\end{keybox}

Release 0.1.0 provides two implemented profiles, \texttt{cstr\_alpha} and \texttt{column\_alpha}. The latter includes D03, an authority-mismatch case in which escalation is required instead of direct manipulation. These profiles demonstrate the evaluation format but do not establish broad industrial generality. The proposed Tennessee Eastman study is therefore presented as a future portability test rather than as a current result.

\section{Audience, Scope, and Decision Context}

\subsection{Who should use this report}
The primary audience is researchers and engineering teams developing or evaluating industrial AI advisory systems. The framework may also help vendors and plant engineering groups define test cases, inspect failure modes, and compare advisory-system configurations under the same plant profile. The resulting evidence may inform assurance review, procurement due diligence, and the design of supervised trials, but release 0.1.0 is not itself a procurement standard, site-acceptance method, or deployment-approval mechanism.

\subsection{Problem statement}

Existing evaluations commonly ask whether a model can classify a fault, explain sensor behaviour, retrieve technical knowledge, or interact with a simulator. These capabilities are relevant, but they do not establish whether a recommendation is acceptable under the operating conditions represented in the test. Industrial advisories are interpreted within a physical and institutional environment that includes safety instrumented functions, operating procedures, alarm-management rules, sensor-quality limitations, role-based authority, and time-dependent process consequences.

The resulting evaluation question is:

\begin{keybox}
\centering
\textbf{Given the available evidence, operating context, authority level, and physical state, is the proposed action admissible under the encoded plant profile, and what evidence supports that verdict?}
\end{keybox}

\claim{Core thesis.} Industrial LLM evaluation should judge the proposed action separately from the diagnosis that precedes it. Answer-level correctness is an insufficient proxy for action admissibility.

\claim{Scope of claim.} \ib\ specifies and implements a reference action-evaluation framework. A verdict states only that one action record passed or failed one versioned plant profile under its encoded evidence, authority, procedure, consequence model, verification horizon, and evaluator configuration. It does not establish that the model, profile, or plant is safe. This separation between evaluation evidence and deployment approval is consistent with the risk-management framing of the NIST AI RMF~\cite{nistairmf}.

In this report, \emph{safety-governed} refers to this explicit eligibility process. It denotes evaluation under documented safety-relevant constraints, not functional-safety certification or a guarantee of safe deployment.

\section{Related Work}

ADMITBench sits at the intersection of four research areas: general evaluation of LLM agents, industrial and prognostics benchmarks, runtime guardrails for tool-using agents, and runtime assurance for cyber--physical systems. Relevant industrial-agent studies, including the authors' prior work, provide application context for action-level evaluation but form only one part of the broader literature.

\subsection{General benchmarks for interactive and consequential agents}

General agent benchmarks established that language models must be evaluated through interaction rather than static question answering. AgentBench evaluates multi-turn reasoning and decision-making across eight environments~\cite{agentbench}, while TheAgentCompany places agents in a simulated workplace where they browse, write and execute code, and communicate with other workers~\cite{agentcompany}. These benchmarks expose long-horizon planning and task-completion failures, but their primary outcome is whether a task is completed rather than whether each proposed intervention is admissible under a domain safety case.

Safety-oriented agent benchmarks move closer to consequential action evaluation. ToolEmu emulates high-stakes tools and scenarios to identify risky agent behaviour at scale~\cite{toolemu}. AgentDojo evaluates tool-using agents under indirect prompt injection and untrusted external data~\cite{agentdojo}. These works demonstrate that nominal task success does not imply safe behaviour. ADMITBench addresses a related but more specific failure mode: an action can be well formed and diagnostically plausible while still failing because its evidence, authority, procedure, reversibility, or predicted physical consequence is inadequate.

\subsection{Industrial and prognostics evaluation}

Process fault detection and diagnosis has a mature model-based literature~\cite{faultdiagnosis}. More recent industrial benchmarks evaluate domain knowledge, maintenance reasoning, and multi-tool workflows. PHM-Bench proposes a multidimensional framework for evaluating large models across prognostics and health-management tasks~\cite{phmbench}. AssetOpsBench evaluates agents performing industrial asset operations and maintenance workflows over heterogeneous data sources~\cite{assetopsbench}. PHMForge further evaluates tool-grounded industrial prognostics through domain-specific Model Context Protocol tools and deterministic execution checks~\cite{phmforge}.

These benchmarks are complementary to ADMITBench. They provide broader coverage of asset-management knowledge, data navigation, prediction, and maintenance orchestration. ADMITBench narrows the evaluation target to the proposed action. Its output reports whether a concrete action record passes checks on record integrity, evidence, hazard understanding, authority and procedure, and predicted physical consequences. The two shipped profiles are deliberately smaller than broad PHM suites because they are intended to make every gate verdict replayable against a versioned safety case.

\subsection{Runtime guardrails and policy enforcement for LLM agents}

A growing body of work places an enforcement layer around the agent rather than relying only on model alignment. AgentSpec provides a domain-specific language for defining triggers, predicates, and runtime enforcement actions~\cite{agentspec}. GuardAgent translates natural-language guard requests into executable checks~\cite{guardagent}, while ShieldAgent constructs verifiable policy representations and checks agent action trajectories against them~\cite{shieldagent}. These systems support the general architectural principle that the generator and the trusted decision mechanism should be separated.

ADMITBench differs mainly in purpose and in the source of its verdict. It is an offline evaluation framework rather than a universal online guardrail. Its hard decision is jointly conditioned on a compiled industrial safety case, temporally valid evidence, hazard interpretation, delegated authority, procedure requirements, reversibility, escalation logic, and simulator-derived consequences. It also uses a non-compensatory eligibility rule: a hard failure in any of T0--T4 produces no aggregate ranking, rather than allowing stronger performance elsewhere to offset the failed gate.

\subsection{Runtime assurance for cyber--physical systems}

The separation between an advanced component and a trusted safety mechanism has a longer history in runtime assurance. Simplex-style architectures permit an advanced, difficult-to-verify controller to operate only while a trusted decision module can maintain safety or transfer control to a verified fallback. The Black-Box Simplex architecture extends this principle to controllers whose internals are unavailable, using runtime checks to preserve system safety~\cite{blackboxsimplex}. This literature provides an important conceptual precedent for treating a high-performing AI component as untrusted at the execution boundary.

ADMITBench is not itself a real-time switching controller and does not inherit the formal safety guarantees of a verified runtime-assurance architecture. Instead, it applies the separation principle to benchmark design: the LLM proposes an action record, while an independent evaluator determines eligibility and produces an auditable explanation of the first failed gate. Dynamic rollout in T4 supplies consequence evidence, but the benchmark remains an evaluation instrument rather than a certified plant protection layer.

\subsection{Industrial LLM control and fault recovery}
Industrial control and fault-recovery systems provide additional application context for action-level evaluation. Validator--reprompter architectures have demonstrated independent checking and corrective feedback for LLM-generated control actions~\cite{vyas2025ifac}. Digital-twin-supported agents have combined plant knowledge with simulation-based verification for process fault handling~\cite{gill2025etfa}, while related work has connected finite-state recovery planning with bounded continuous control~\cite{vyas2025autonomous} and linked fault detection to knowledge-grounded recovery, deterministic checks, dynamic feasibility validation, and bounded fallback behaviour~\cite{vyas2026ftc}.

These systems illustrate how candidate industrial actions can be generated, corrected, and checked within particular implementations. ADMITBench addresses the complementary question of how those proposed actions can be represented and evaluated using a common, inspectable format. The present release demonstrates this format in two profiles; broader comparison across independently developed plants and vendors remains to be established.

\section{Why Diagnosis-Only Evaluation Is Insufficient}

A model may identify a cooling-loss fault in an exothermic reactor and still recommend an intervention that violates a procedure, relies on a suspect flow measurement, exceeds its delegated authority, or arrives after the process has crossed a safety boundary. Diagnosis addresses what may be happening. Operational decision support must also specify what is proposed, whether the available evidence supports it, who is permitted to act, and whether the action remains acceptable over the relevant process horizon.

Figure~\ref{fig:diagnosis-action} separates diagnosis quality from action admissibility.

\begin{figure}[htbp]
\centering
\begin{tikzpicture}[font=\footnotesize\sffamily]

\def\cw{6.1}
\def\ch{2.15}

\draw[fill=red!8, draw=admitred] (0,\ch) rectangle (\cw,2*\ch);
\draw[fill=green!8, draw=admitgreen] (\cw,\ch) rectangle (2*\cw,2*\ch);
\draw[fill=orange!10, draw=orange!65!black] (0,0) rectangle (\cw,\ch);
\draw[fill=blue!7, draw=admitblue] (\cw,0) rectangle (2*\cw,\ch);

\node[align=center, text width=5.4cm] at (0.5*\cw,1.5*\ch) {\textbf{Hidden action failure}\\Correct diagnosis; unsafe or inadmissible action};
\node[align=center, text width=5.4cm] at (1.5*\cw,1.5*\ch) {\textbf{Desired recovery}\\Correct diagnosis; admissible action};
\node[align=center, text width=5.4cm] at (0.5*\cw,0.5*\ch) {\textbf{Full failure}\\Wrong diagnosis; inadmissible action};
\node[align=center, text width=5.4cm] at (1.5*\cw,0.5*\ch) {\textbf{Safe containment}\\Incomplete diagnosis; appropriate hold, verify, or escalate};

\node[above] at (0.5*\cw,2*\ch) {ACTION INADMISSIBLE};
\node[above] at (1.5*\cw,2*\ch) {ACTION ADMISSIBLE};
\node[rotate=90, align=center, text width=2.5cm] at (-0.90,1.5*\ch) {DIAGNOSIS\\CORRECT};
\node[rotate=90, align=center, text width=2.5cm] at (-0.90,0.5*\ch) {DIAGNOSIS INCOMPLETE OR WRONG};

\draw[thick, draw=admitgray]
    (-0.10,2*\ch) -- (-0.34,2*\ch) -- (-0.34,0) -- (-0.10,0);
\draw[thick, draw=admitgray] (-0.34,\ch) -- (-0.10,\ch);

\end{tikzpicture}
\caption{Diagnosis and action quality are separate. \ib\ is designed to expose a correct diagnosis paired with an inadmissible action and to recognise conditionally correct containment under uncertainty.}
\label{fig:diagnosis-action}
\end{figure}

This distinction is operationally important. A plausible answer can fail because it skipped a mandatory step, acted outside authority, used untrusted measurements, or proposed an irreversible intervention without adequate evidence. Conversely, a recommendation to hold, verify, or escalate may be the appropriate response when evidence is incomplete or corrupted.

\section{Method Overview}

\subsection{Evaluation boundary}

\begin{figure}[htbp]
\centering
\begin{tikzpicture}[
    node distance=1.0cm and 0.85cm,
    stage/.style={draw=admitblue, thick, rounded corners=2pt, fill=admitlight, align=center, minimum height=1.35cm, text width=4.0cm, font=\small\sffamily},
    resource/.style={draw=black!55, rounded corners=2pt, fill=black!3, align=center, minimum height=1.35cm, text width=4.0cm, font=\small\sffamily},
    output/.style={draw=admitgreen, thick, rounded corners=2pt, fill=green!5, align=center, minimum height=1.35cm, text width=4.0cm, font=\small\sffamily},
    flow/.style={-{Latex[length=2.2mm]}, thick, draw=black!70},
    support/.style={-{Latex[length=2.0mm]}, semithick, dashed, draw=admitgray}
]
\node[stage] (context) {\textbf{1. Scenario packet}\\Plant state, alarms, and trust flags\\Procedure phase and authority};
\node[stage, right=of context] (model) {\textbf{2. Model interaction}\\LLM advisory\\under evaluation};
\node[stage, right=of model] (record) {\textbf{3. Action record}\\Parsed action and evidence\\Confidence and recovery path};

\node[resource, below=of context] (contract) {\textbf{Versioned contract}\\Parser schema and action grammar\\Released cartridge};
\node[stage, below=of model] (verifier) {\textbf{4. Evaluation checks}\\T0--T4 eligibility\\T5 utility ranking; T6 audit};
\node[output, below=of record] (report) {\textbf{5. Evaluation report}\\Eligibility and failed checks\\Audit trace; eligible-only score};

\draw[flow] (context) -- (model);
\draw[flow] (model) -- (record);
\draw[flow] (record.south west) -- (verifier.north east);
\draw[flow] (contract) -- (verifier);
\draw[flow] (verifier) -- (report);
\draw[support] (contract) -- (context);
\end{tikzpicture}
\caption{Evaluation boundary. The upper row captures the advisory-system interaction; the lower row applies the versioned schema, plant profile, eligibility checks, utility ranking, and audit checks. The output is an inspectable evaluation report.}
\label{fig:evaluation-loop}
\end{figure}
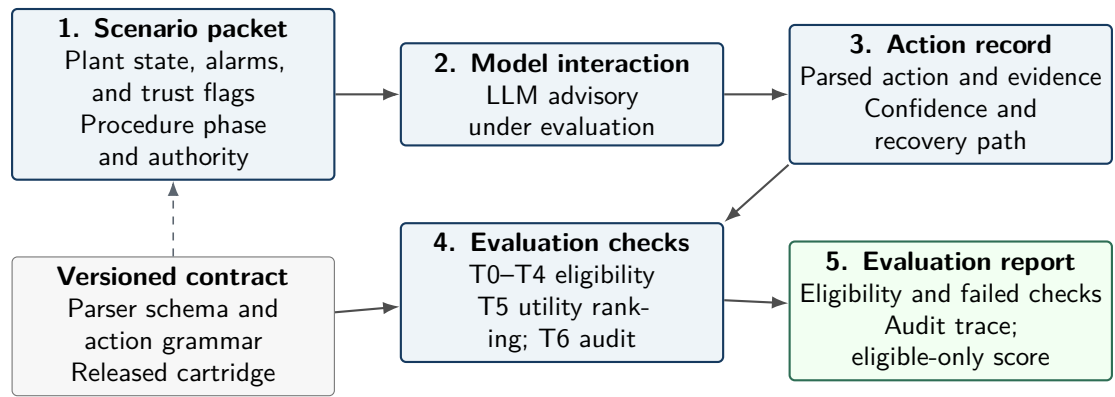

\ib\ evaluates the path from plant context to a structured advisory record and then to rule- and model-based checks. Figure~\ref{fig:evaluation-loop} shows the evaluation boundary.

\subsection{Action record as the unit of evaluation}

The action record contains the information required to evaluate a recommendation as an operational proposal rather than as free text. At minimum, it records:

\begin{enumerate}
    \item the proposed advisory action and action parameters;
    \item the asserted diagnosis or fault class;
    \item cited evidence and source identifiers;
    \item required checks or verification steps performed;
    \item authority scope and deployment configuration;
    \item confidence or uncertainty information;
    \item reversibility, fallback, and recovery information; and
    \item traceability, source linkage, and audit fields.
\end{enumerate}

The record is evaluated for T0--T4 eligibility before any aggregate is considered. A record that fails a hard gate receives \texttt{aggregate=None}; it is reported as ineligible rather than placed on the same ranking scale as eligible actions.

\subsection{Plant-profile architecture}

The implementation separates three elements:

\begin{itemize}
    \item \textbf{Common evaluation logic:} the checks and reporting rules intended to remain unchanged across plant profiles.
    \item \textbf{Reusable schema:} the structure used to encode authority, safety, procedures, evidence, actions, and outcomes.
    \item \textbf{Plant-specific content:} the manifest, system graph, safety-case graph, procedure cases, and any consequence-verification resources used by the checks.
\end{itemize}

The repository calls each versioned plant profile a \emph{cartridge}. The purpose of this separation is practical: adding a fired heater, compressor anti-surge system, or Tennessee Eastman process model~\cite{tep} should primarily require new plant data, actions, procedures, and parameters. If a new profile requires undocumented plant-specific logic in the common evaluator, the claimed separation has not been maintained.
The four files that constitute a release 0.1.0 plant profile are summarised in Table~\ref{tab:cartridge}.
\begin{table}[htbp]
\centering
\caption{Cartridge files shipped in ADMITBench release 0.1.0.}
\label{tab:cartridge}
\small
\begin{tabularx}{\textwidth}{@{}>{\raggedright\arraybackslash}p{0.27\textwidth}Y@{}}
\toprule
\textbf{Artifact} & \textbf{Role in the released cartridge} \\
\midrule
\texttt{manifest.yaml} & Declares profile metadata and binds the cartridge resources used by the evaluator. \\
\texttt{system\_graph.jsonl} & Encodes the operational system graph, states, transitions, and action-relevant structure. \\
\texttt{safety\_case\_graph.jsonl} & Encodes the safety-case graph used for evidence, hazard, admissibility, and consequence checks. \\
\texttt{procedures\_cases.jsonl} & Encodes the evaluated procedure and case records, including profile-specific operating and escalation conditions. \\
\bottomrule
\end{tabularx}
\end{table}

The audit trace is generated by the evaluation harness; it is not a fifth cartridge file. Release 0.1.0 does not record stochastic generation seeds, so repeated-generation provenance must be documented externally when model sampling is used.

\section{Authority Model}

The same recommendation can be acceptable or unacceptable depending on the authority granted to the system. \ib\ distinguishes three deployment configurations:

\begin{itemize}
    \item \textbf{A0 -- Advisory:} the model may explain, prioritise, recommend verification, and suggest escalation, but cannot issue direct plant manipulation as an executable instruction.
    \item \textbf{A1 -- Supervised:} the model may propose bounded corrective actions for human approval when procedural preconditions are satisfied.
    \item \textbf{A2 -- Constrained autonomous:} the model may propose actions within a predefined grammar and operating envelope, while safety gates and authority limits remain hard boundaries.
\end{itemize}

Authority may narrow during an episode because of degraded evidence, abnormal operating conditions, or site policy. It cannot expand because the model claims that a broader action is necessary. A technically useful recommendation still fails the authority check when it exceeds the role assigned to the system.

\section{Evaluation Hierarchy}
\label{sec:evaluation-hierarchy}

T0--T4 are non-compensatory eligibility gates. T5 is used only to rank utility among records that have passed every hard gate. T6 records traceability and audit evidence and does not alter eligibility. Table~\ref{tab:tiers} defines the concern, evaluation question, and role associated with each tier.

\begin{table}[t]
\centering
\caption{ADMITBench tier hierarchy in repository release 0.1.0.}
\label{tab:tiers}
\scriptsize
\begin{tabularx}{0.98\textwidth}{@{}>{\raggedright\arraybackslash}p{0.055\textwidth}>{\raggedright\arraybackslash}p{0.18\textwidth}>{\raggedright\arraybackslash}p{0.25\textwidth}Y>{\raggedright\arraybackslash}p{0.11\textwidth}@{}}
\toprule
\textbf{Tier} & \textbf{Name} & \textbf{Primary concern} & \textbf{Evaluation question} & \textbf{Role} \\
\midrule
T0 & Record integrity and safety-case compilation & Required fields, types, identifiers, graph references, and successful compilation of the record against the selected safety case & Is the action record structurally complete, internally valid, and compilable against the declared profile? & Hard gate \\
T1 & Evidence and state validity & Evidence availability, provenance, temporal validity, state consistency, and support for the declared decision context & Is the decision based on evidence that is available, valid, and consistent with the represented plant state? & Hard gate \\
T2 & Hazard and causal understanding & Recognition of the decision-relevant hazard, sufficient causal understanding to justify the selected action, and appropriate escalation when that understanding is unavailable& Does the record address the decision-relevant hazard sufficiently to justify the selected action, or appropriately hand the decision upward? & Hard gate \\
T3 & Action admissibility & Authority, permitted action family, procedure conformance, preconditions, reversibility, recovery, and escalation obligations & Is the proposed action admissible for this actor, state, procedure, and evidence condition? & Hard gate \\
T4 & Physical consequence verification & Dynamic or rule-based consequence checks, safety-envelope boundaries, response horizon, and unsafe-delay conditions & Does the proposed action avoid unacceptable physical consequences over the declared verification horizon? & Hard gate \\
T5 & Utility ranking among eligible actions & Relative usefulness, efficiency, disruption, or preference among otherwise eligible actions & Among actions that pass T0--T4, which eligible action is preferable? & Ranking only \\
T6 & Traceability and audit & Source linkage, gate explanations, record lineage, and reproducibility of the deterministic judgement & Can a reviewer reconstruct why the record passed, failed, or was ranked as it was? & Audit only \\
\bottomrule
\end{tabularx}
\end{table}
For T4, the minimum verification horizon and any unsafe-delay condition are fixed by the versioned plant profile or evaluator configuration; the evaluated advisory system cannot shorten the horizon to exclude delayed consequences. The hierarchy is non-compensatory: a failure at any of T0--T4 makes the record ineligible, and later tiers cannot offset it. T5 is considered only after eligibility has been established. T6 preserves the information needed to replay the implemented checks. Reports should distinguish the first failed tier from the complete set of failed checks, because a record may contain several independent defects.

The hierarchy is standards-aware but does not replace the engineering standards that govern functional safety, safety instrumented systems, and alarm management~\cite{iec61511,iec61508,isa18}.

\section{Eligibility, Rule-Based Evaluation, and Reporting}

\subsection{Eligibility before ranking}

Another language model may be useful for qualitative analysis, but it is not used as the primary judge in T0--T4. These tiers are executed using declared record checks, profile rules, and consequence models. The execution is deterministic for a fixed record, evaluator version, profile, and simulator configuration; the engineering judgements encoded in the profile remain human-authored and must be reviewed. A failed hard gate produces no aggregate score. Accordingly, determinism here means reproducible execution of declared rules; it does not imply that the encoded engineering judgements are universal, objective, or sufficient for a real plant.

\Needspace{8\baselineskip}
Let $h_t(e) \in \{0,1\}$ indicate whether episode $e$ hard-fails tier $t$, for $t\in\{0,1,2,3,4\}$. Define

\begin{equation}
H(e)=\max_{t\in\{0,1,2,3,4\}} h_t(e).
\end{equation}

The aggregate is then

\begin{equation}
A(e)=
\begin{cases}
\varnothing, & H(e)=1,\\[0.35em]
0.15s_1(e)+0.15s_2(e)+0.20s_3(e)+0.30s_4(e)+0.20s_5(e), & H(e)=0
\end{cases}.
\end{equation}

For $t\in\{1,2,3,4,5\}$, $s_t(e)\in[0,1]$ is the deterministic normalised subscore associated with tier T$t$. Let $C_t(e)$ denote the set of applicable scored checks for that tier, let $c_{tj}(e)\in[0,1]$ be the outcome of check $j\in C_t(e)$, and let $w_{tj}\geq 0$ be its versioned cartridge weight. The tier subscore is

\begin{equation}
s_t(e)=\frac{\sum_{j\in C_t(e)}w_{tj}c_{tj}(e)}{\sum_{j\in C_t(e)}w_{tj}}.
\end{equation}

Here, $s_1$ measures evidence and state quality, $s_2$ hazard and causal understanding, $s_3$ action admissibility quality, $s_4$ verified physical-consequence quality and safety margin, and $s_5$ relative utility among eligible actions. The applicable checks, their deterministic outcome rules, and their weights must be declared by the versioned evaluator or cartridge; an empty or invalid check set is a configuration error rather than an implicit zero. For T1--T4, these subscores describe quality only within the eligible region and cannot compensate for a hard-gate failure. T5 is therefore ``ranking only'' in the eligibility hierarchy even though $s_5$ contributes to the aggregate after eligibility has been established.

The coefficients in Equation~(2) are release-specific reporting weights, not universal statements about the relative importance of the tiers. Evaluations should publish the underlying tier vector and, when the aggregate is used for comparison, report sensitivity to plausible alternative weights. The aggregate should therefore be treated as a secondary summary over eligible records rather than as the primary benchmark verdict.

In Equation~(2), $\varnothing$ corresponds to the repository value \texttt{aggregate=None}. T0 is an integrity and compilation gate and is not assigned a compensating weight. T6 is reported separately as traceability and audit evidence.

Assigning \texttt{None} rather than zero keeps ineligible records outside the ranking of eligible actions. The primary report should therefore preserve the full gate report, the first failed tier, the complete set of failed checks, and the audit trace rather than presenting the aggregate alone.

\subsection{Primary report object}

For every record, the evaluator should expose at least

\begin{equation}
\mathbf{v}(e)=[T0,T1,T2,T3,T4,T5,T6,A(e)],
\end{equation}

where each tier retains its verdict and supporting evidence, $A(e)$ is defined only for eligible records, and the report includes both the first failed tier and the complete set of failed checks. Population-level calibration, repeated-generation consistency, and robustness analyses may be added as separate statistics, but they do not alter the hard-gate result.

\subsection{Selective risk and abstention}

An advisory system should not be penalised for withholding a corrective instruction when the evidence or authority is insufficient. \ib\ therefore treats \emph{verify}, \emph{hold}, and \emph{escalate} as legitimate actions when those responses are required by the plant profile.

\subsection{Recommended report outputs}

A technical evaluation report should include at least:

\begin{itemize}
    \item pass/fail and first-failure counts by tier;
    \item diagnosis accuracy separated from action admissibility;
    \item action-family and authority-scope errors;
    \item evidence-degradation and profile-specific challenge-case performance;
    \item confidence coverage, calibration, and selective-risk curves where the model supplies confidence;
    \item parser and trace-completeness rates;
    \item dynamic safety-envelope outcomes and horizon declarations;
    \item reproducibility metadata, release and profile identifiers, version hashes, and simulator configuration; and
    \item representative failure traces that can be independently inspected.
\end{itemize}

\begin{keybox}[Interpretation principle]
A hard-gate failure makes the record ineligible; it is not a low score that can be compensated by utility or explanation quality. Results should be read as a profile of observed failures under the released cases, not as a general claim that a model is ``safe.''
\end{keybox}

\section{Implemented Plant Profiles}

Release 0.1.0 ships two implemented plant profiles, referred to as cartridges in the repository. They use the same evaluator and tier semantics while supplying different system and safety-case content, as summarised in Table~\ref{tab:implemented-cartridges}.

\begin{table}[htbp]
\centering
\caption{Implemented cartridges in ADMITBench release 0.1.0.}
\label{tab:implemented-cartridges}
\small
\begin{tabularx}{\textwidth}{@{}>{\raggedright\arraybackslash}p{0.23\textwidth}>{\raggedright\arraybackslash}p{0.25\textwidth}Y@{}}
\toprule
\textbf{Profile} & \textbf{Process setting} & \textbf{Role in the release} \\
\midrule
\texttt{cstr\_alpha} & Jacket-cooled continuous stirred-tank reactor & Provides a compact continuous-process profile for evaluating evidence, hazard understanding, action admissibility, and physical consequence verification. It also connects to the authors' earlier CSTR fault-recovery work~\cite{vyas2026ftc}, while shifting the emphasis from controller performance to record-level admissibility and auditability. \\
\texttt{column\_alpha} & Distillation-column operations & Provides a second implemented profile and includes D03, an authority-mismatch case requiring escalation rather than direct action, making differences in delegated authority and escalation obligations observable. \\
\bottomrule
\end{tabularx}
\end{table}

The two profiles show that the same evaluator can be populated for more than one process setting. They do not establish broad industrial generality: both were authored within the same project, and neither provides independent evidence that the profile boundary will remain stable for substantially different plants.

\section{Evaluation Protocol}

\subsection{Model interaction}

Each advisory-system configuration receives a standardised scenario containing plant state, alarms, sensor-trust flags, procedural context, and authority level. The configuration includes the model, system prompt, available tools or retrieved information, action grammar, parser, sampling settings, and any reprompting policy. Its output is converted into a structured advisory record containing the proposed action, diagnosis, confidence, cited evidence, verification steps, escalation decision, and rationale fields.

Unparseable responses should be reported separately. Parser compatibility is not merely a formatting concern: an advisory that cannot be reliably interpreted cannot be reliably governed or audited.

The T4 verification horizon is governed by the profile-level rule defined in Section~\ref{sec:evaluation-hierarchy}. The selected horizon should be justified using the process dynamics and the time available for detection, intervention, and recovery.

\subsection{Current case organisation and future splits}

Release 0.1.0 does not implement formal public, held-out, and adversarial splits. Its evaluated cases are supplied through \texttt{procedures\_cases.jsonl} within each cartridge. Accordingly, results from the current release should be described as evaluations over the released case suite, not as held-out generalisation evidence.

For future benchmark releases, three explicitly versioned populations would strengthen the evaluation design:

\begin{itemize}
    \item \textbf{Public development population:} supports implementation, debugging, and reproducibility.
    \item \textbf{Held-out evaluation population:} tests unseen but in-distribution operating cases under documented contamination controls.
    \item \textbf{Adversarial population:} targets conflicting or stale evidence, near-boundary trajectories, procedural ambiguity, and authority mismatch.
\end{itemize}

These are recommendations for later releases, not features claimed for repository version 0.1.0.

\subsection{Evaluation questions}

The protocol is designed to answer four practical questions:

\begin{enumerate}
    \item Do models that diagnose correctly also recommend admissible actions?
    \item Do models preserve admissible behaviour when evidence is degraded or conflicting?
    \item Do particular profiles or challenge cases expose substantially higher hard-gate failure rates?
    \item Does trace quality remain distinct from hard-gate eligibility and utility ranking?
\end{enumerate}

The methodology deliberately separates the evaluation contract from any particular model leaderboard. Empirical claims should identify the exact released case suite and profiles; claims about held-out generalisation require future splits and contamination controls that are not present in release 0.1.0.

\section{Failure Modes the Method Is Designed to Reveal}

\subsection{Diagnosis-action decoupling and first-failure localisation}

A system can produce the right diagnosis but the wrong action, or an appropriate containment action without identifying the precise root cause. T2 should therefore require the minimum hazard understanding needed to justify the selected action, rather than complete causal diagnosis in every case. \ib\ separates T2 hazard understanding, T3 action admissibility, and T4 physical consequence verification. The report records both the first failed tier and all failed checks so that the imposed gate order does not hide additional defects.

\subsection{Calibration under uncertainty}

Confidence affects human interpretation. A highly confident wrong advisory can contribute to automation bias, while an underconfident correct advisory may be ignored. Calibration, coverage, and selective risk should therefore be treated as first-class outputs rather than optional model-quality statistics.

\subsection{Challenge-case and profile concentration}

Aggregate eligible-record scores can conceal where hard failures concentrate. Reporting first-failure counts by case and profile reveals whether authority mismatch, evidence degradation, or unsafe physical consequences are concentrated in a small number of operating situations. This release supports case- and profile-level stratification; a formal adversarial split remains future work.

\subsection{Trace plausibility versus correctness}

Explanations can sound credible even when the proposed action is unsafe. Rationale is valuable for debugging, operator review, and procurement assessment, but it remains auxiliary. It cannot override safety, procedure, sensor-trust, or dynamic-outcome checks.

\section{Interpretation of Evaluation Results}

\subsection{Why a vector is more useful than a leaderboard scalar}

Two eligible advisory-system configurations can have the same aggregate score while failing in different ways. One may be poorly calibrated under degraded evidence; another may frequently select a less useful but still admissible action. The tier vector, full failed-check distribution, profile stratification, and eligible-record utility score preserve distinctions that a single aggregate can hide.

\subsection{Why provenance matters}

Evidence should be trusted because of its source and provenance, not because the model describes it as trustworthy. The harness should distinguish trusted simulator or plant channels, retrieved documents, user-entered context, stale values, and unsupported model assertions. Source trust is a provenance property rather than a natural-language property.

\subsection{Where language models fit}

\ib\ does not assume that the language model replaces a fault tree, procedure engine, safety system, or controller. Structured engineering artifacts remain the source of truth. This separation is consistent with the authors' earlier architectures, in which LLM agents propose candidate actions while digital twins, state-machine structure, interlocks, and deterministic validators constrain or reject them~\cite{vyas2025ifac,gill2025etfa,vyas2025autonomous,vyas2026ftc}. The model is evaluated as a layer that integrates partial evidence, operating context, natural-language information, and competing constraints into a proposed advisory. The evaluation asks whether the resulting transition is admissible under uncertainty.

\subsection{How decision-makers should use the evidence}

The evidence can support comparison of advisory-system configurations under the same plant profile, identification of recurring action or evidence failures, and selection of cases for further engineering or operator evaluation. At the present stage, these results are best treated as technical evaluation evidence. Their use in procurement, supervised trials, or deployment decisions requires additional site-specific review and independently validated plant profiles.

\section{Portability Validation Plan}
\label{sec:tep-demonstrator}

A stronger test of the plant-profile abstraction is to move from the released CSTR profile to the Tennessee Eastman (TE) plant-wide process-control problem~\cite{tep}. TE introduces interacting units, recycle, multiple manipulated and measured variables, coupled fault propagation, and a larger operating context. It therefore provides a useful test of whether plant-specific content can change without requiring undocumented changes to the common evaluator.

Because TE implementations can differ in numerical behaviour, the demonstrator should use a frozen, tested simulator revision and publish solver, initialisation, and termination details~\cite{tep_revision}.

\claim{Portability criterion.} The action-record schema, tier meanings, parser contract, audit-trace structure, and eligibility-before-ranking rule should remain unchanged. TE-specific equipment, actions, procedures, safety predicates, sensor-trust profiles, scenarios, and dynamic parameters should enter through the new plant profile. Any required change to the common evaluator should be documented and treated as evidence against the current portability claim.
The evidence required to evaluate this portability claim is summarised in Table~\ref{tab:tep-portability}.
\begin{table}[t]
\centering
\caption{Pre-registered evidence for a Tennessee Eastman portability demonstrator. No TE results are claimed in this report.}
\label{tab:tep-portability}
\scriptsize
\begin{tabularx}{\textwidth}{@{}>{\raggedright\arraybackslash}p{0.14\textwidth}>{\raggedright\arraybackslash}p{0.24\textwidth}>{\raggedright\arraybackslash}p{0.28\textwidth}Y@{}}
\toprule
\textbf{Dimension} & \textbf{Held fixed} & \textbf{TE cartridge contribution} & \textbf{Evidence required} \\
\midrule
Interface & Action-record fields, parser failure policy, and authority semantics & TE equipment identifiers, bounded action grammar, units, and operating modes & Schema-conformance tests pass without evaluator-specific exceptions. \\
Knowledge & T0--T6 meanings and eligibility-before-ranking semantics & System graph, safety-case graph, procedure cases, and profile manifest & Every verdict resolves to a versioned cartridge source and deterministic check. \\
Dynamics & Rollout API, horizon declaration, and safe-envelope contract & Frozen TE model, parameters, solver settings, initial states, and termination rules & Repeated rollouts reproduce within declared numerical tolerances. \\
Cases & Released case-record contract & Coupled disturbances, recycle effects, stale or conflicting evidence, and authority mismatch & Case provenance and hashes are reported; any future development, held-out, or adversarial split is declared explicitly. \\
Reporting & Gate report, first-failure tier, generated audit trace, and \texttt{aggregate=None} rule & TE-specific strata and failure taxonomy & Reports retain the same structure and disclose cartridge-authoring effort and all scoring-engine changes. \\
\bottomrule
\end{tabularx}
\end{table}

This demonstrator would test whether the method extends from a compact thermal-hazard example to plant-wide interactions. The claim remains prospective until the frozen profile, conformance tests, authoring effort, evaluator changes, and results are released.

\section{Implementation Workflow}

A practical implementation can be organised as follows.

\begin{enumerate}
    \item \textbf{Define the decision boundary.} Specify whether the model is explanatory, advisory, supervised, or permitted to propose constrained actions.
    \item \textbf{Freeze the action-record contract.} Define required fields, allowed actions, parser failure policy, evidence identifiers, and lineage requirements.
    \item \textbf{Author and review the cartridge.} Populate the released four-file cartridge contract: manifest, system graph, safety-case graph, and procedure cases, with controlled engineering review.
    \item \textbf{Validate the deterministic checks.} Test each tier independently and publish conformance tests for the schema and scoring implementation.
    \item \textbf{Run repeated evaluations.} Separate model-generation variability from deterministic replay of a fixed action record; because release 0.1.0 does not capture generation seeds, retain that provenance in the external run configuration.
    \item \textbf{Review failure traces.} Use scenario-stratified failures to tighten authority, prompt design, parser constraints, or operating scope.
    \item \textbf{Conduct site-specific assurance.} Treat benchmark evidence as input to operator studies, management of change, safety review, and controlled trials rather than as replacement for them.
\end{enumerate}

\section{Limitations and Non-Claims}

\ib\ evaluates advisory records against the rules and models encoded in a selected plant profile. Its outputs do not replace HAZOP, LOPA, SIL verification, alarm rationalisation, operator training, management of change, or functional-safety assessment, and they do not authorise deployment.

The two released cartridges are deliberately limited. The CSTR and distillation-column profiles are tractable reference settings, not complete representations of industrial process operations. Evidence from two same-project cartridges does not establish broad industrial generality. Generality must be demonstrated incrementally through additional cartridges, independent review, and documented portability tests.

The quality of the result depends directly on the quality of the plant profile. Incomplete procedures, unreviewed cause-and-effect entries, incorrect safety constraints, or unsuitable consequence models can produce misleading verdicts. Profile authoring should therefore be a controlled, human-reviewed engineering activity, and the source and rationale for each rule should be traceable.

Operator effects are only partially captured. Alarm fatigue, automation bias, trust erosion, situation awareness, and tacit local knowledge require empirical operator studies and site-specific review~\cite{endsley1995}. \ib\ measures technical preconditions for usable advisory behaviour; it does not prove human-factors adequacy in a real control room.

\section{Next Development Steps}

\subsection{Technical milestones}

\begin{itemize}
    \item stabilise and independently review the released \texttt{cstr\_alpha} and \texttt{column\_alpha} cartridges;
    \item introduce explicit development, held-out, and adversarial splits in a future release;
    \item publish a stable parser and action-record contract;
    \item release schema and evaluator conformance tests;
    \item report verified reference evaluations across a small set of frontier and open-weight models;
    \item execute the pre-registered Tennessee Eastman portability demonstrator;
    \item introduce independent scoring and cartridge-review governance; and
    \item develop a registry for versioned, human-reviewed cartridges.
\end{itemize}

\subsection{Scientific milestones}

\begin{itemize}
    \item quantify the diagnosis-to-admissible-action gap;
    \item test whether calibration predicts safe abstention under sensor uncertainty;
    \item evaluate challenge-case and future adversarial-split failure concentration as deployment-relevant statistics;
    \item compare LLM advisories against operator, rule-based, and conservative baselines; and
    \item formalise tacit-knowledge capture as a structured review workflow.
\end{itemize}

\section{Conclusion}

\textbf{In industrial operations, the answer is not the unit of safety; the proposed action is.} A correct diagnosis does not guarantee that the corresponding recommendation is supported by valid evidence, permitted under delegated authority, procedurally admissible, or physically adequate.

\ib\ makes this gap explicit through a structured action record, versioned plant profiles, non-compensatory T0--T4 eligibility gates, T5 utility ranking among eligible actions, and T6 audit evidence. Release 0.1.0 demonstrates this evaluation format through a common evaluator and two same-project profiles. It does not yet establish broad industrial generality, independent profile portability, human-factors adequacy, or deployment readiness. Those claims require independently authored profiles, broader case populations, reference evaluations, sensitivity studies, and site-specific assurance.

The present contribution is therefore a reference framework and public implementation for making action-level evaluation more explicit, inspectable, and replayable. It is neither a safety certificate nor a replacement for accountable engineering judgement.

\section*{Ethics and Safety Statement}
\addcontentsline{toc}{section}{Ethics and Safety Statement}

This work is intended to improve technical evaluation of industrial AI advisory systems. The released scenarios are benchmark cases rather than operating instructions. Any deployment requires site-specific engineering and safety review, validated plant models and procedures, management of change, operator training, and accountable approval.

\end{document}